\documentclass[referee,sn-mathphys-num]{sn-jnl}

\usepackage{graphicx}%
\usepackage{multirow}%
\usepackage{amsmath,amssymb,amsfonts}%
\usepackage{amsthm}%
\usepackage{mathrsfs}%
\usepackage[title]{appendix}%
\usepackage{xcolor}%
\usepackage{textcomp}%
\usepackage{manyfoot}%
\usepackage{booktabs}%
\usepackage{algorithm}%
\usepackage{algorithmicx}%
\usepackage{algpseudocode}%
\usepackage{listings}%
\usepackage{subcaption} %
\usepackage{adjustbox} %
\usepackage{colortbl} %
\usepackage{xcolor} %
\usepackage{hhline} %

\theoremstyle{thmstyleone}%
\theoremstyle{thmstyletwo}%

\theoremstyle{thmstylethree}%

\begin{document}

\title[Evaluating and improving crop-yield forecasting methods during extreme drought]{Evaluating and improving crop-yield forecasting methods during extreme drought}


\author*[1]{\fnm{Shrey} \sur{Gupta}}\email{shrey.gupta.skg@gmail.com}

\author[1]{\fnm{Yi} \sur{Ming}}\email{Yi.Ming@bc.edu}

\author[1]{\fnm{George} \sur{Mohler}}\email{mohlerg@bc.edu}

\affil[1]{\orgname{Boston College}, \country{US}}





\abstract{The impact of climate variability on food production has led to the creation of various forecasting models that uses machine learning (ML), numerical weather predictors (NWP) or a hybrid of ML-NWP models to identify structural and physical relationships between meteorological drivers and crop growth, in order to predict crop yield.
Droughts, for example the 2012 Midwestern US (Corn Belt) drought, are extreme events that affect crop production and test the limits of  these forecasting models.  
Using 16 meteorological drivers as predictors, we compare ML (non-deep learning) and deep learning forecasting models to predict the county-level corn yield for the extreme drought year, 2012. 
This forecasting problem is characterized by a dissimilarity between the feature distributions of the training and test data, where the meteorological conditions of the extreme drought year fall outside the range of historically observed values.
Additionally, the dataset consists of spatial and temporal irregularities where counties with missing yields introduce spatial sparsity and the use of only a subset of daily values per year introduce temporal sparsity.
To overcome this, we use sample weighting and feature selection as modifications to improve our forecasting models.
These modifications lead to an improvement for ML models; however, the deep learning model VITA shows little to no improvement. 
While VITA outperforms the ML models with or without modifications, our current study sheds light on the effect of dissimilarity between train and test feature distributions on forecasting models, compares deep learning versus non-deep learning models, and introduces modifications that are effective for non-deep learning models.}




\maketitle

\section{Introduction}~\label{sec:introduction}
Climate variability (natural fluctuations in climate) and anthropogenic climate change (long-term shifts driven by human activity) has influenced the global agricultural productivity by altering the frequency and intensity of extreme events such as droughts and floods~\cite{ortiz2021anthropogenic,yang2024climate,shi2025extreme}.
Among these extremes events, droughts are considered highly consequential as they affect (reduce) the soil moisture, surface energy and hydrological cycles.
This directly impacts the crop yield with cascading economic and societal consequences~\cite{lobell2014greater}.
The 2012 year drought that took place in the Midwestern United States (US) (also called the Corn Belt) was one of the most severe droughts witnessed by the region in the last 50 years.
It affected over $80\%$ of the contiguous United States and reduced the national corn yields for that year by approximately $13\%$~\cite{boyer2013us}.

A range of approaches exist for crop yield forecasting that include statistical learning, machine learning (ML), and deep learning (DL) models\cite{you2022incorporating,lemercier2021distribution,you2017deep,he2023physics,liu2023task,liu2022machine,wang2018deep,ansarifar2021interaction}.
We consider ML and DL models as two independent sets where the former does not use neural networks.
These ML and DL models struggle to forecast extreme events (droughts, floods, and more) as the data consist of very few representative samples.
This under representation is caused due to (1) a dissimilarity between the feature distributions of the training and test data, where the meteorological conditions of the extreme drought year fall outside the range of historically observed values, and (2) {\em Data irregularities}, a scenarios where missing spatiotemporal data affects the model's performance. 
For conceptual shorthand, we refer to this dissimilarity between train and test feature distributions as {\em feature distribution mismatch} throughout this study.
{\bf Therefore, the goal of this study is to evaluate and improve the performance of yield forecasting models during extreme drought by considering it a feature distribution mismatch and data irregularity problem.}

Much of the recent progress in crop yield forecasting has been driven by using large neural networks and rich training datasets with the assumption that performance improves under scaling laws, i.e., model and dataset size~\cite{kaplan2020scaling}, even for extreme events.
While solutions based on scaling improves the forecasting model's performance, they remain a black box as they are unable to explain the underlying agroclimatic processes that affect crop yield.
Therefore, {\bf we answer the following three questions in this work: (1) How do ML and DL models perform under feature distribution mismatch and data irregularity constraints and does the model complexity impact generalization (prediction performance)? (2) 
Are meteorological drivers in the data enough to capture the variations in the crop yield and which drivers are most useful in capturing these variations? (3) Do explainable AI techniques further improve the model performance?}

Our study addresses the problem of accurately estimating annual county-level corn yield under the constraints of feature distribution mismatch and data irregularities and utilizing only meteorological drivers to analyze the performance of the forecasting models. 
The data consists of annual county-level corn yield combined with daily meteorological drivers (features) across the Midwestern US.
The data irregularities emerge due to inconsistency in (1) {\em spatial coverage} (spatial irregularities) from missing county-level yield data, and (2) {\em temporal coverage} (temporal irregularities) as daily measurements do not span the full year.
We improve the model performance by using sample weighing and selecting important features derived using the SHAP analysis.

In Section~\ref{sec:background}, we discuss the relevant background and related work in this problem space.
Then we discuss the model and modifications employed in Section~\ref{sec:methodology} and their forecasting results in Section~\ref{sec:evaluation}.
Our study's takeaway and future work is discussed in Section~\ref{sec:discussion}.

\section{Background and Related Works}~\label{sec:background}
Crop yield is a challenging prediction problem, as yields are affected by a confluence of meteorological drivers, land-use patterns, and crop management practices where each vary across space and time. 
Meteorological drivers influence yield indirectly compared to the dependent factors such as soil properties, crop physiology, and management practices~\cite{petrovic2026analyzing}.
However, these meteorological drivers alone are able to explain greater than one-third of yield variability~\cite{knight2024impact,ray2015climate,teasdale2017meteorological}, which is a motivation to study the impact of these variations on the crop yield when observations are in-situ and aggregated over a large area (US county).
Previous studies have introduced ML and DL frameworks that forecast the 2012 drought in the Midwestern US. 
In the following sections, we discuss the various yield forecasting models from the literature as well as the datasets used -- gridMET and NASA power.




    
    
    

\begin{figure}[t]
    \centering
    \begin{subfigure}[t]{0.50\linewidth}
        \centering
        \includegraphics[width=\linewidth]{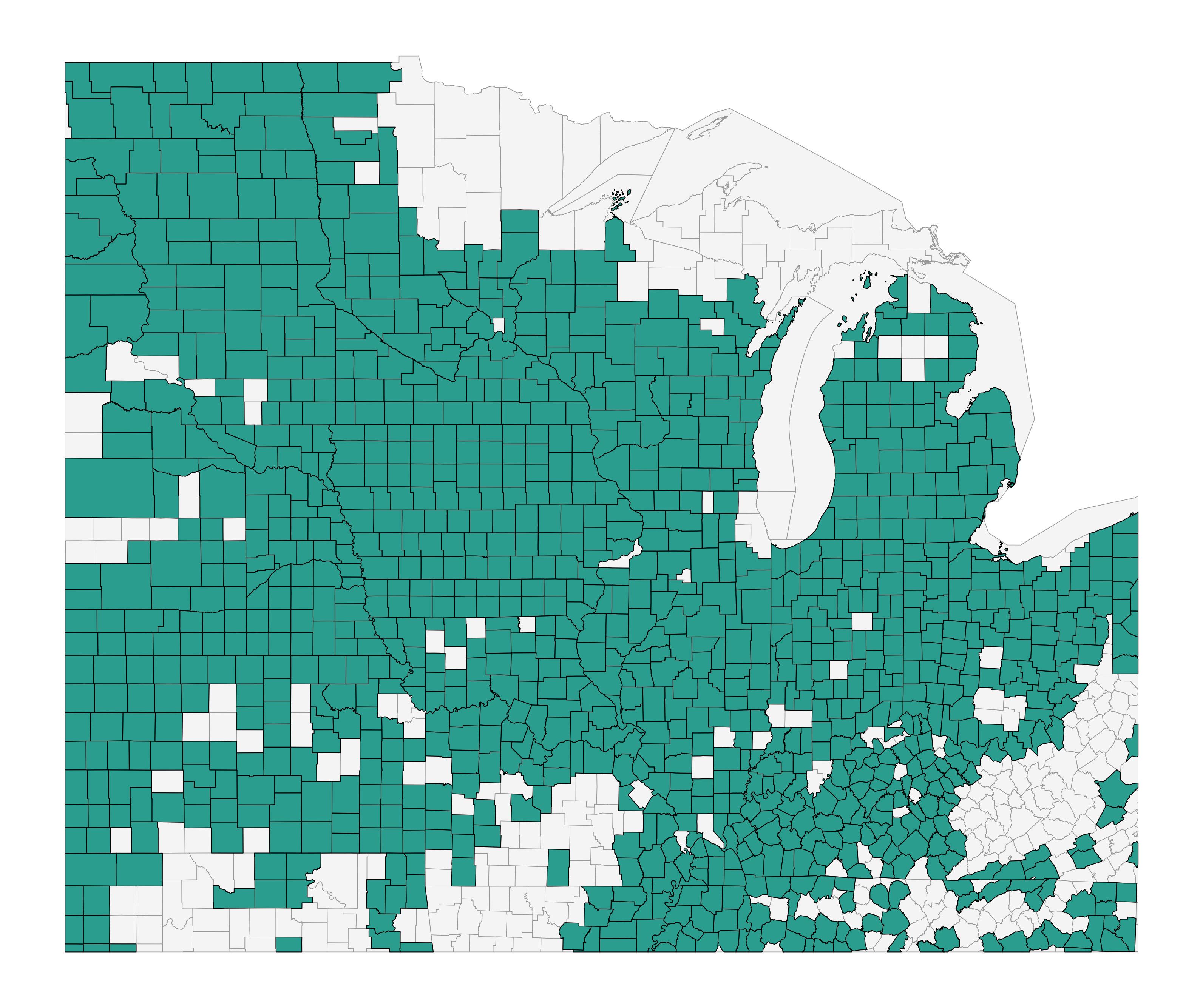}
        \caption{Gridmet data coverage}
        \label{fig:gridmet}
    \end{subfigure}%
    \hfill
    \begin{subfigure}[t]{0.46\linewidth}
        \centering
        \includegraphics[width=\linewidth]{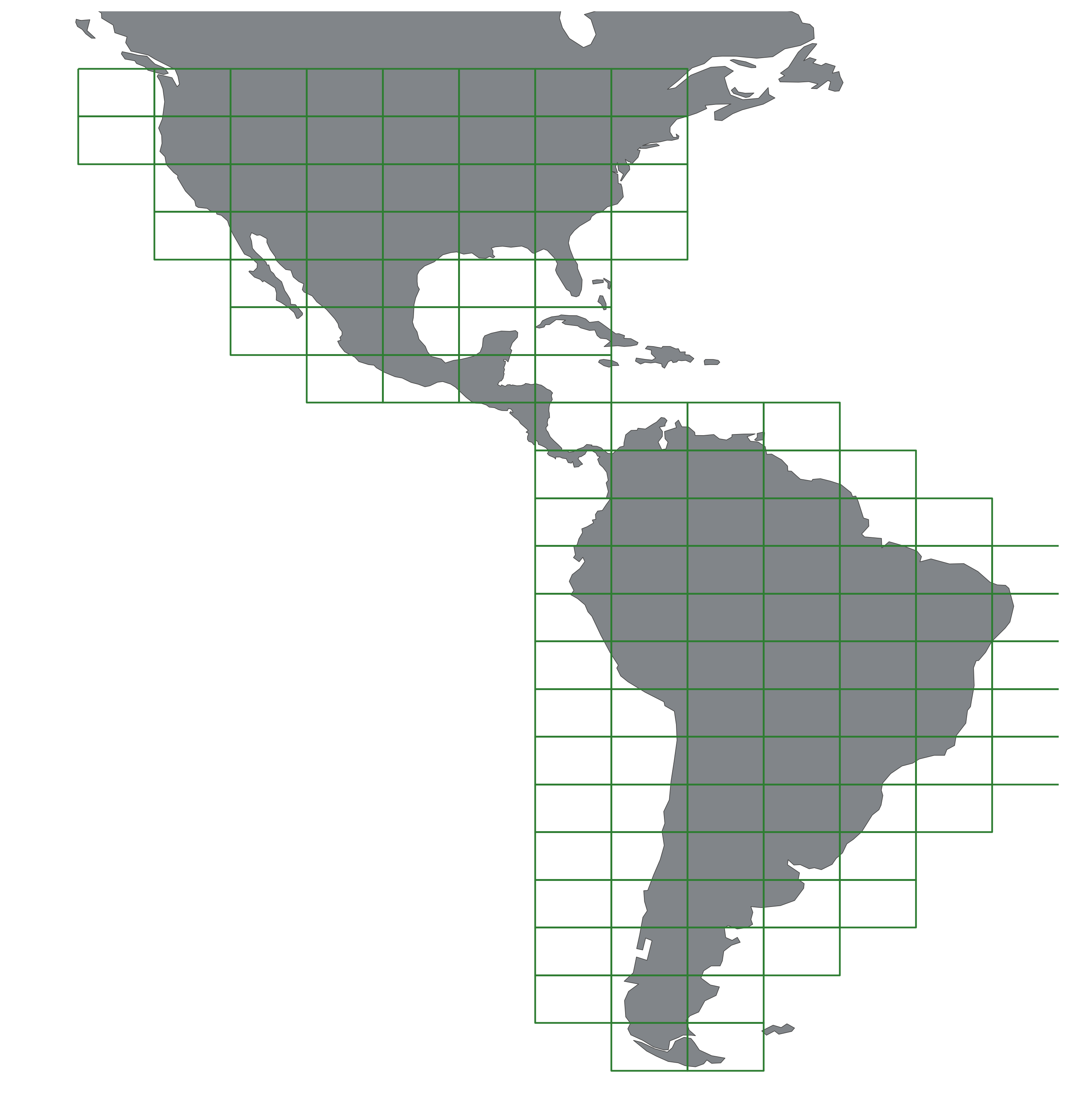}
        \caption{NASA Power data}
        \label{fig:nasa-power}
    \end{subfigure}
    \caption{Region of coverage for (a) Gridmet data (US Cornbelt) with counties where grey are the total number of counties and green are the available county data from the year 2012, and (b) NASA Power data (North and South America) with 5x8 grids.}
    \label{fig:training-data}
\end{figure}

\paragraph{{\bf gridMET Dataset}} 
The gridMET dataset~\cite{abatzoglou2013development,liu2022machine} spans the US Corn belt, that is geographically located between latitudes 35.58° to 49.58° and longitudes 80.8° to 102.8°, and includes 1,413 counties as shown in Figure~\ref{fig:gridmet}.
The dataset spans 1980 to 2018 and consist of two types of data irregularities: spatial irregularities due to missing yields across certain counties and years, and temporal irregularities, as the daily measurements only cover the crop phases, sowing, growing, and harvesting phases, than the complete year.
The county-level corn yields were obtained from the USDA National Agricultural Statistics Service and the daily meteorological data of the gridMET dataset was collected and processed by Abatzoglou et al.~\cite{abatzoglou2013gridmet} at a 1/24\textdegree{} ($\sim$4 km) resolution. 
The dataset includes 16 meteorological drivers (features), namely specific humidity, mean vapor pressure deficit, precipitation, minimum and maximum relative humidity, downwelling solar radiation, minimum and maximum air temperature, wind speed and direction, reference grass and alfalfa evapotranspiration, energy release component, burning index, and 100-hour and 1000-hour dead fuel moisture (FM100 and FM1000).
The crop phases cover 210 days ranging from 10 April to 6 November per year.

\paragraph{{\bf Nasa Power Dataset}}
The NASA Power dataset, generated using the NASA Power API by~\citet{hasan2025vita}, contains 39 years (1984-2022) of data at 0.5\textdegree{} spatial resolution.
It is used by the VITA (Variational Inference Transformer for Asymmetric data) model~\cite{hasan2025vita} as a pretraining dataset.
Whereas, the gridMET dataset is used for finetuning and forecasting.
The NASA Power dataset consists of 116 grid tiles (160 geo-referenced points per tile) spanning the Americas as shown in Figure~\ref{fig:nasa-power}. 
It includes 31 meteorological drivers measured at daily granularity, normalized to a sequence length of 365 per year, with missing values backward filling and leap year days omitted. 
These daily sequences are then aggregated to weekly measurements and standardized using the mean and standard deviation of the continental US, which generates approximately 100K sequences, that are used for pretraining the model.

\paragraph{{\bf Yield Forecasting}} 
There exist numerous studies that focus on crop yield forecasting using various ML models~\cite{you2022incorporating,lemercier2021distribution,you2017deep,fan2022gnn,he2023physics,liu2023task,liu2022machine,wang2018deep,ansarifar2021interaction}.
Therefore, a critical open question is then determining models, trained on meteorological features, that are useful forecasters during extreme droughts.
It should be noted that the data used in these studies usually contain high-resolution satellite or in-situ data or a combination of both.
Within the in-situ data, there are both meteorological and land-use drivers, such as soil moisture, that directly impact the crop yield.
For this problem, we focus on the transfer learning model, {\em Tradaboost.R2}, extreme-value model, {\em Extreme Lasso}, and the deep learning model, {\em VITA}~\cite{hasan2025vita}.
These models show promise given the drought scenario when only meteorological drivers are present, and the dataset consists of spatiotemporal irregularities.
In the following sections, we elaborate on the studies that use forecasting models for yield prediction during extreme droughts.

\subsection{Crop yield forecasting models}
Zhong et al.~\cite{zhong2023detect} provides a multi-task learning framework that embeds attention in a LSTM architecture to capture seasonality and periods of extreme stress during the crop growing phases.
Their model shows a strong yield forecasting performance where the drought indicator (derived feature), {\em Killing Degree Days}, which measures heat stress experienced by crop, is the indicative feature for variations in the yield.
However, their dataset consists of meteorological, soil, and vegetation index (NDVI) drivers that improves their model's performance in capturing high spatiotemporal variability. 
Xu et al.~\cite{xu2021machine} utilize a similar heterogeneous and rich dataset with meteorological, soil, and management practice drivers, to forecast corn yield.
They determine {\em vapor pressure deficit} (VPD) and temperature as major meteorological drivers of yield variability and soil's water capacity and organic matter content as land drivers.
Shahhosseini et al.~\cite{shahhosseini2021coupling} show that combining ML models with process-based models can improve the accuracy of crop yield prediction.
They do this by using outputs of these process-based models like DSSAT~\cite{hoogenboom2019dssat} as inputs to the ML models.
The process-based model output is calculated using important indicators such as soil-water balance, phenology, and canopy development.
While the above models have a strong forecasting performance, they utilize a rich heterogeneous dataset, compared to our study that focuses on answering yield variability captured by meteorological drivers for extreme events.

\subsection{Drought forecasting models}
Drought indicators are derived features that quantify the onset, and intensity of drought conditions.
Several studies focus on forecasting these indicators rather than yield or soil moisture.
Brust et al.~\cite{brust2021droughtcast} introduce Droughtcast, an LSTM model, to predict Evaporative Stress Index (ESI), for forecasting flash droughts.
The authors use meteorological drivers and vegetation indices to predict upcoming changes in drought conditions.
Their model effectively captures the rapid transitions that indicate flash droughts by learning temporal patterns indicative of the decline in Evaporative Stress Index (ESI).
Similarly, Deangelis et al.~\cite{deangelis2020prediction}, evaluate forecasting performance for the Subseasonal Experiment (SubX) models to capture the rapid intensification of the 2012 US drought. 
The SubX models showed low forecasting performance for spring and early-summer precipitation deficits and moderate forecasting performance for temperature anomalies, leading to a failure in anticipating the final drought patterns.
While these studies demonstrate the use of drought indicators such as ESI (remote-sensing based), NDVI (vegetation based), and more, for capturing land-atmosphere dynamics, our setting is intentionally constrained to meteorological drivers.


\section{Methodology}~\label{sec:methodology}
We first introduce the models we use for our comparative study along with the feature engineering and model modifications we use in combination with the existing models. 

\subsection{Prediction Models}
For our experiments, we consider a roster of machine learning models and the deep learning model, VITA. 
Our goal here is to compare various forecasting models that use kernels, trees, heuristics to overcome extreme-values, and transfer learning.
VITA (Variational Inference Transformer for Assymetric data) is a more recent DL model that forecasts crop yield. 
It has shown promise to identify and capture similar drought patterns in the dataset, i.e. temporal sequences of meteorological drivers, similar to the test (drought) year, using its variational inference and attention architecture.

The ML models we use differ in how they learn the relationship between the features and the labels where the features (meteorological drivers) are standardized before modeling. 
For a given feature vector $\mathbf{x}_i$, the model prediction is denoted as $\hat{y}_i = f(\mathbf{x}_i)$ and the corresponding residual is $r_i = y_i - f(\mathbf{x}_i)$, where $y_i$ is the observed target value. 
A model is denoted using the functional form, $f(\cdot)$, and the loss function is defined over the residuals ($r_i$) between the model outputs ($f(\mathbf{x}_i)$) and the observed ground-truth ($y_i$).

\paragraph{{\bf Machine learning models:}} 
For the machine learning models, we assume a linear relationship between the features and the target, such that the model output is $f(\mathbf{x}_i) = \mathbf{x}_i^\top \boldsymbol{\beta}$, where $\boldsymbol{\beta}$ are the coefficients.
We use approaches such as {\bf Ridge} and {\bf Lasso}, which apply regularization (L2 and L1, respectively) for generalization. 
In Ridge regression, the coefficients $\boldsymbol{\beta}$ are calculated by minimizing the L2 regularized squared–error loss,
\[
\mathcal{L}_{\text{Ridge}}(\boldsymbol{\beta})=\frac{1}{n}\sum_{i=1}^{n}(y_i-\mathbf{x}_i^\top\boldsymbol{\beta})^2+\lambda\lVert\boldsymbol{\beta}\rVert_2^2
\]

Whereas, in Lasso regression, the coefficients $\boldsymbol{\beta}$ are calculated by minimizing the L1 regularized squared–error loss,
\[
\mathcal{L}_{\text{Lasso}}(\boldsymbol{\beta})=\frac{1}{n}\sum_{i=1}^{n}(y_i-\mathbf{x}_i^\top\boldsymbol{\beta})^2+\lambda\lVert\boldsymbol{\beta}\rVert_1.
\]
where $n$ is the total number of training samples, $\lambda$ is the regularization parameter that controls the penalty applied to the coefficients, $\lVert \boldsymbol{\beta} \rVert_2^2$ and $\lVert \boldsymbol{\beta} \rVert_1$ are the squared $\ell_2$ and $\ell_1$ norm respectively.

We use tree-based models, {\bf Random Forest} and {\bf Gradient Boosting regressor}, which do not assume a linear relationship for $f(\cdot)$, but instead recursively partition the feature space to learn predictions.
Each tree in the Random Forest minimizes the mean-squared error,
$\mathcal{L}_{\text{tree}} =
\frac{1}{n_T}
\sum_{i \in T}
\left(y_i - \hat{y}_i\right)^2,
$
where a single decision tree $T(\cdot)$ is denoted by $\hat{y}_i = T(\mathbf{x}_i)$, and $n_T$ denotes the number of samples in the node.
The Random Forest model has $M$ trees where the ensemble prediction is defined as,
$f_{\text{RF}}(\mathbf{x}_i) =
\frac{1}{M}
\sum_{m=1}^{M}
T_m(\mathbf{x}_i)$

We use a kernel-based model, \textbf{Support Vector Regressor (SVR)}, which uses kernel function to operate in a reproducing kernel Hilbert space. 
We use a linear kernel $k(\mathbf{x}_i, \mathbf{x}_j) = \mathbf{x}_i^\top \mathbf{x}_j$, such that the prediction function is $f(\mathbf{x}_i) = \sum_{j=1}^{n} w_j\, k(\mathbf{x}_j, \mathbf{x}_i) + b = \mathbf{w}^\top \mathbf{x}_i + b$.
\[
\mathcal{L}_{\text{SVR}}(\beta)
=
\frac{1}{2}\lVert \mathbf{w} \rVert_2^2
+
C \sum_{i=1}^{n}
L_\epsilon\bigl(y_i - f(\mathbf{x}_i)\bigr),
\]
where $L_\epsilon(r) = \max\bigl(0,\, |r| - \epsilon\bigr)$ is the $\epsilon$-insensitive loss, $C > 0$ is the regularization parameter, and $\epsilon > 0$ controls the insensitivity, such that |r| is the error.

We use extreme-value model, {\bf Extreme Lasso}~\cite{chang2021sparse} and {\bf Huber regression}~\cite{huber1992robust}. 
Let $r_i = y_i - \mathbf{x}_i^\top \boldsymbol{\beta}$ be the residuals (error) and $\delta > 0$ be the threshold (also known as Huber threshold) for the Huber model. The loss for a single residual in the model is defined as,
\[
\rho_\delta(r) =
\begin{cases}
\frac{1}{2} r^2, & \text{if } |r| \le \delta,\\[4pt]
\delta \bigl(|r| - \tfrac{1}{2}\delta\bigr), & \text{if } |r| > \delta,
\end{cases}
\]
and the loss function for Huber regression is defined as
\[
\mathcal{L}_{\text{Huber}}(\boldsymbol{\beta})
=
\frac{1}{n}
\sum_{i=1}^{n}
\rho_\delta\bigl(y_i - \mathbf{x}_i^\top \boldsymbol{\beta}\bigr).
\]

Similarly for the Extreme Lasso model, $\gamma > 2$ represents the weight given to the extreme residuals. The loss function for the model can then be written as,
\[
\mathcal{L}_{\text{ExtremeLasso}}(\boldsymbol{\beta})
=
\frac{1}{n}
\sum_{i=1}^{n}
\bigl\lvert y_i - \mathbf{x}_i^\top \boldsymbol{\beta} \bigr\rvert^{\gamma}
+
\lambda \lVert \boldsymbol{\beta} \rVert_1,
\]
where $\lambda > 0$ controls the $L_1$ regularization strength.

Transfer learning models are known to handle feature distribution mismatch observed when the test data is a different domain. 
We use {\bf TrAdaBoost.R2}, a boosting-based regression transfer approach that combines the source data (domain to learn from; large sample set) and target data (domain to predict upon; fewer sample set) and reweights source samples that are dissimilar to the target samples~\cite{pardoe2010boosting}. 
If the source and target domains are,
$\mathcal{D}_S = \{(x_i^S, y_i^S)\}_{i=1}^{n_S}$ and 
$\mathcal{D}_T = \{(x_j^T, y_j^T)\}_{j=1}^{n_T}$, 
where $w_{S,i}^{(t)}$ and $w_{T,j}^{(t)}$ denote the weights at each boosting iteration $t$.
At iteration $t$, TrAdaBoost.R2 learns a base regressor $h_t : \mathbb{R}^d \to \mathbb{R}$, similar to the  prediction function $f(\cdot)$, by minimizing the weighted normalized absolute error as,
\[
\mathcal{L}^{(t)}(\beta)
=
\sum_{i=1}^{n_S} w_{S,i}^{(t)}\, e_{S,i}^{(t)}
+
\sum_{j=1}^{n_T} w_{T,j}^{(t)}\, e_{T,j}^{(t)},
\]
where $e_{S,i}^{(t)} = \frac{\bigl|y_i^S - h(x_i^S)\bigr|}{D^{(t)}}$, $e_{T,j}^{(t)} = \frac{\bigl|y_j^T - h(x_j^T)\bigr|}{D^{(t)}}$ and $D^{(t)} = \max_{k}\bigl|y_k - h(x_k)\bigr|$.
After training the $h_t$ learner, the weights are updated differently for target and source points to focus on high residual target samples and dissimilar source samples. 
For a normalization constant $Z_t$ and a transfer parameter $\beta \in (0,1)$, the updates are
\[
w_{T,j}^{(t+1)}
=
\frac{w_{T,j}^{(t)}\, \beta^{-e_{T,j}^{(t)}}}{Z_t},
\qquad
w_{S,i}^{(t+1)}
=
\frac{w_{S,i}^{(t)}\, \beta^{\,e_{S,i}^{(t)}}}{Z_t}.
\]
Hence, the source samples with larger residuals $e_{S,i}^{(t)}$ are exponentially down-weighted over iterations, while the target samples are up-weighted.

\paragraph{{\bf Deep learning model:}}
We use VITA~\cite{hasan2025vita}, which is a decoder-free variational transformer model, which learns latent weather representations from the satellite data during pretraining and transfers the learned weather representation to a sparse subset of features at finetuning.
The yield for a sample $i$ is represented as
$
\hat{y}_i = f_{\text{VITA}}(\mathbf{x}_i;\theta),
$
where 
$\mathbf{x}_i$ denotes the meteorological drivers and $\theta$ represents the neural network parameters to be learned.

The input for the pretraining stage consists of a weekly time-series,
$\mathbf{x} = \{\mathbf{x}_t\}_{t=1}^{T}$, such that $\mathbf{x}_t \in \mathbb{R}^{34},$ where the number of meteorological drivers is 31 along with spatial and temporal features (latitude, longitude, and year), and $T=364$ denotes the sequence length. 
At each timestep $t$, the input feature vector $\mathbf{x}_t \in \mathbb{R}^{34}$ is first mapped to a higher-dimensional embedding space via a linear projection, $\mathbf{e}_t = W\mathbf{x}_t + b$, where $W \in \mathbb{R}^{200 \times 34}$ and $b \in \mathbb{R}^{200}$.
To preserve the temporal order of the sequence, a sinusoidal positional encoding $\mathbf{p}_t \in \mathbb{R}^{200}$ is added to the embedding as, $\tilde{\mathbf{e}}_t = \mathbf{e}_t + \mathbf{p}_t$, which allows the Transformer encoder to recognize different timesteps.
Next, the embedded sequence $\{\tilde{\mathbf{e}}_t\}_{t=1}^T$ is processed by a Transformer encoder with four layers.`
Each layer consists of multi-headed self-attention (10 heads) and a position-wise feedforward network of width 800 with GELU activations and residual connections, that generates the contextual representation, 
$\mathbf{h}_t \in \mathbb{R}^{200}$.

The contextual representations $\mathbf{h}_t \in \mathbb{R}^{200}$ produced by the Transformer encoder summarizes the meteorological conditions at each time step $t$, and also accounts for temporal dependencies present in the sequence. 
These contextual representations are then mapped into a latent space that captures the underlying meteorological conditions in a probabilistic manner.
VITA achieves this by assuming that at each time step $t$, the latent weather state $\mathbf{z}_t \in \mathbb{R}^{31}$ follows a Gaussian variational posterior distribution conditioned on the input sequence,
\[
q_{\boldsymbol{\phi}}(\mathbf{z}_t \mid \mathbf{x}_t)
=
\mathcal{N}
\bigl(
\boldsymbol{\mu}_t,\,
\operatorname{diag}(\boldsymbol{\sigma}_t^2)
\bigr),
\]
where $\boldsymbol{\phi}$ denotes the parameters of the variational inference network. 
The mean vector $\boldsymbol{\mu}_t$ represents the inferred latent weather state at time $t$, while the diagonal covariance $\operatorname{diag}(\boldsymbol{\sigma}_t^2)$ quantifies uncertainty associated with this inference.
The parameters $\boldsymbol{\mu}_t$ and $\log \boldsymbol{\sigma}_t^2$ are obtained via a linear projection of the Transformer output $\mathbf{h}_t$ from $\mathbb{R}^{200}$ to $\mathbb{R}^{62}$, split equally into mean and log-variance. 
The variance is enforced to be positive through the exponential transformation,
\[
\boldsymbol{\sigma}_t^2 = \exp\!\left(\log \boldsymbol{\sigma}_t^2\right).
\]
This variational formulation allows the model to represent meteorological conditions as distributions rather than deterministic embeddings, which allows to capture uncertainty in the sequences.

During the pretraining stage, VITA is trained in a self-supervised manner using only meteorological drivers as inputs. 
A subset of these drivers is randomly masked across all timesteps, and the model reconstructs the masked values from the latent variables $\mathbf{z} = \{\mathbf{z}_t\}_{t=1}^T$. 
This forces the latent space to capture coherent temporal and cross-feature dependencies.
The pretraining loss function is defined as
\[
\mathcal{L}_{\text{pretrain}}
=
- \log p(\mathbf{x}_{\text{masked}} \mid \mathbf{z})
+
\alpha\,
\mathrm{KL}
\!\left(
q_{\boldsymbol{\phi}}(\mathbf{z} \mid \mathbf{x})
\,\|\, 
p(\mathbf{z})
\right),
\]
where the first term corresponds to a Gaussian negative log-likelihood over the masked entries.
The second term is the Kullback--Leibler divergence between the variational posterior and the seasonality-aware prior, weighted by $\alpha$, which regularizes the latent representations and prevents deviation from physically plausible seasonal patterns.

For yield prediction, the pretrained Transformer encoder is reused to infer latent states $\{\mathbf{z}_t\}_{t=1}^T$. These states are aggregated using a learned temporal attention mechanism. First, scalar attention scores are computed as
\[
s_t = \operatorname{MLP}(\mathbf{z}_t),
\]
where MLP is the Multi-layer Perceptron. This is followed by normalizing attention weights,
$\alpha_t=\frac{\exp(s_t)}{\sum_{k=1}^{T} \exp(s_k)}.$
The aggregated latent representation is then
\[
\bar{\mathbf{z}}
=
\sum_{t=1}^{T}
\alpha_t \mathbf{z}_t,
\qquad
\bar{\mathbf{z}} \in \mathbb{R}^{31}.
\]

The attention MLP has architecture $31 \rightarrow 16 \rightarrow 1$ with GELU activation, producing a scalar weight $\alpha_t$ per timestep. 
The latent vector $\bar{\mathbf{z}} = \sum_t \alpha_t \mathbf{z}_t \in \mathbb{R}^{31}$ is concatenated with $n_{\text{past}}$ historical yield values, capturing soil and management effects. 
This combined representation is passed through a yield prediction MLP with architecture,
\[
(31 + n_{\text{past}}) \rightarrow 120 \rightarrow 1
\]
with GELU activation to produce the final yield, $\hat{y}_i$. During fine-tuning, the reconstruction term is disabled, and the model is optimized using a mean-squared error loss with a KL regularization term retained to preserve the pretrained latent structure,
\[
\mathcal{L}_{\text{finetune}}
=
\frac{1}{n}
\sum_{i=1}^{n}
\left(y_i - \hat{y}_i\right)^2
+
\beta\,
\mathrm{KL}
\!\left(
q_{\boldsymbol{\phi}}(\mathbf{z} \mid \mathbf{x})
\,\|\, 
p(\mathbf{z})
\right).
\]

\subsection{Data and model adaptation}\label{sec:data-model-adaptation}
We describe the preprocessing steps applied to the gridMET dataset 
for regression modeling, the explainability analysis used for feature 
selection, and the sample reweighting applied to extreme event years. 
We apply these collectively across our models and refer to them as 
data and model adaptation.

\paragraph{{\bf Feature Engineering}}
The gridMET dataset consists of 210 days of meteorological observations (drivers) per county per year.
For the ML models, we aggregate the daily observations into a rolling 3-month seasonal window spanning from April to October.
For each window, we compute summary statistics such as the window's mean, standard deviation, minimum \& maximum values, quartiles, skewness, and kurtosis capturing the distribution's variance. 
We also expand the yield feature using its historical values to introduce features such as previous 5-year yield mean as well as minimum, maximum, mean, standard deviation and trend for the all the previous yield years. 
We avoid time-series detrending methodologies such as windowed averaging, linear regression, and more. 
\citet{dessavre2019problem} argue that detrending either over-smooths (reduces variance) or under-smooths (increases oscillations) the time-series, and consequently destroying the structure of the series.
Instead we use the trend feature that captures the change (increase/decrease) in yield for each county.

\paragraph{{\bf Explainability analysis}}
\begin{figure}
    \centering
    \begin{subfigure}[t]{0.48\linewidth}
        \centering
        \includegraphics[width=\linewidth]{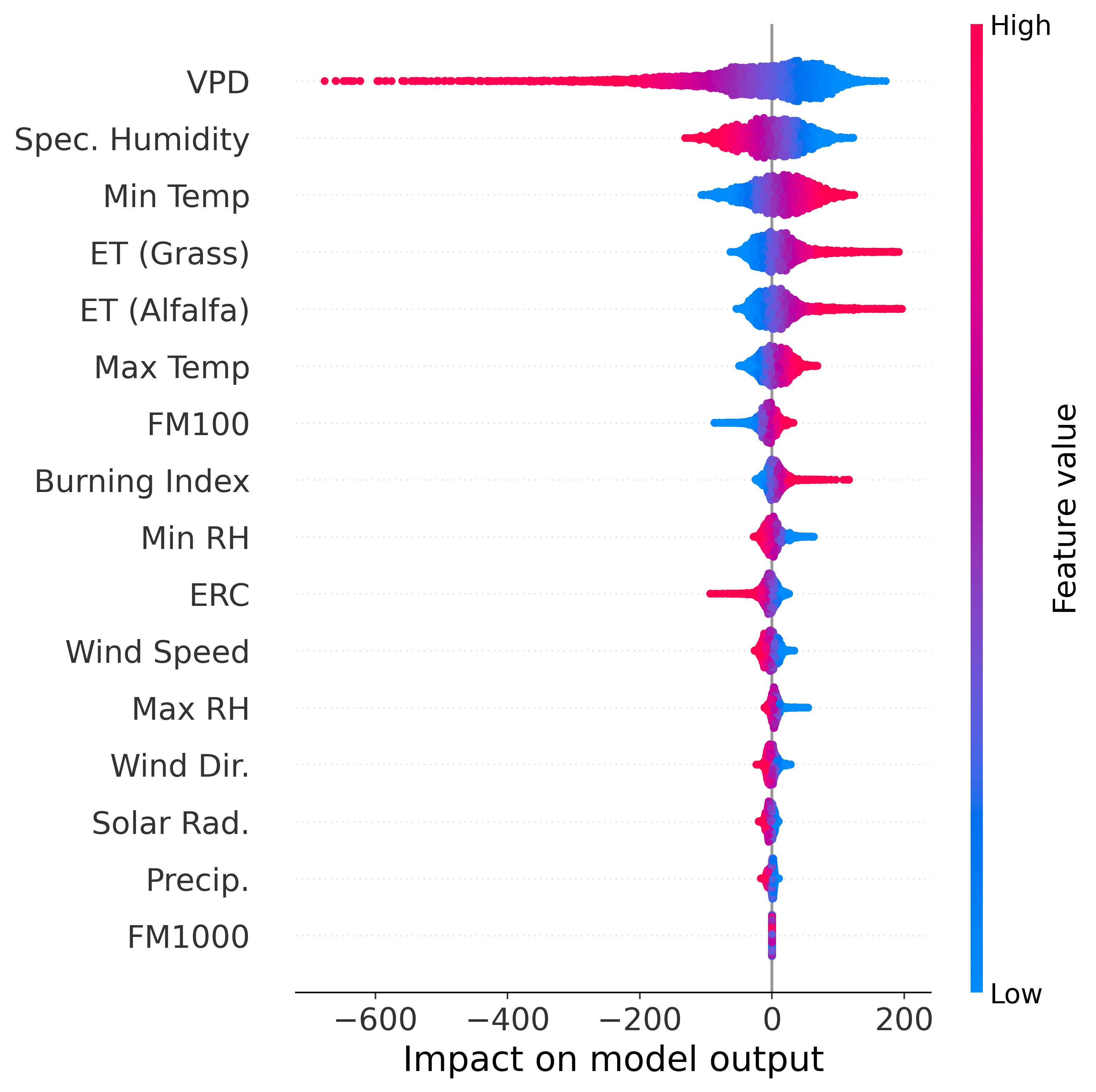}
        \caption{SHAP:2012.}
        \label{fig:shap-2012}
    \end{subfigure}
    \hfill
    \begin{subfigure}[t]{0.48\linewidth}
        \centering
        \includegraphics[width=\linewidth]{Figures/shap_summary_2012.png}
        \caption{SHAP:2013.}
        \label{fig:shap-2013}
    \end{subfigure}
    \caption{SHAP analysis for the test years (a) 2007 to 2011 
    trained on historical years before 2007 and (b) 2008 to 2012 
    trained on historical years before 2008. Impact on model output 
    (x-axis) equates to yield (bushels/acre).}
    \label{fig:shap-analysis}
\end{figure}

SHAP (SHapley Additive exPlanations)~\cite{lundberg2017unified} is a model-agnostic explainability methodology that uses game theory to identify essential features for forecasting.
It computes Shapley values, which represent the average marginal contribution of each feature to the model's predictions across all possible feature combinations, thereby determining feature importance.
We first aggregate the features annually (across the 210 days) and  subsequently train a Lasso regression model on samples before 2007, testing on the window from 2007 to 2011.
While we aim not to use the extreme event year (2012) directly, the 5-year test window allows us to identify short-term meteorological drivers that influence yield forecasting.
Figure~\ref{fig:shap-analysis} shows the SHAP summary, which identifies {\em vapor pressure deficit} (VPD) and {\em evapotranspiration} (ETR) as the most influential meteorological drivers for yield forecasting during 2007 to 2011, while other drivers show little to no influence.

\paragraph{{\bf Extreme event years weighting}}
We additionally identify $4$ previous extreme event years in the U.S. Corn Belt -- 1983, 1988, 1993, and 2002 -- that substantially impacted crop production.
These years witnessed high moisture deficits and heat stress during the growing season that resulted in significant yield reductions.
We apply weighting and attention to the samples of these years in our models to improve forecasting.

\paragraph{{\bf Model adaptation}}
\begin{figure*}[t]
    \centering
    \includegraphics[width=\textwidth]{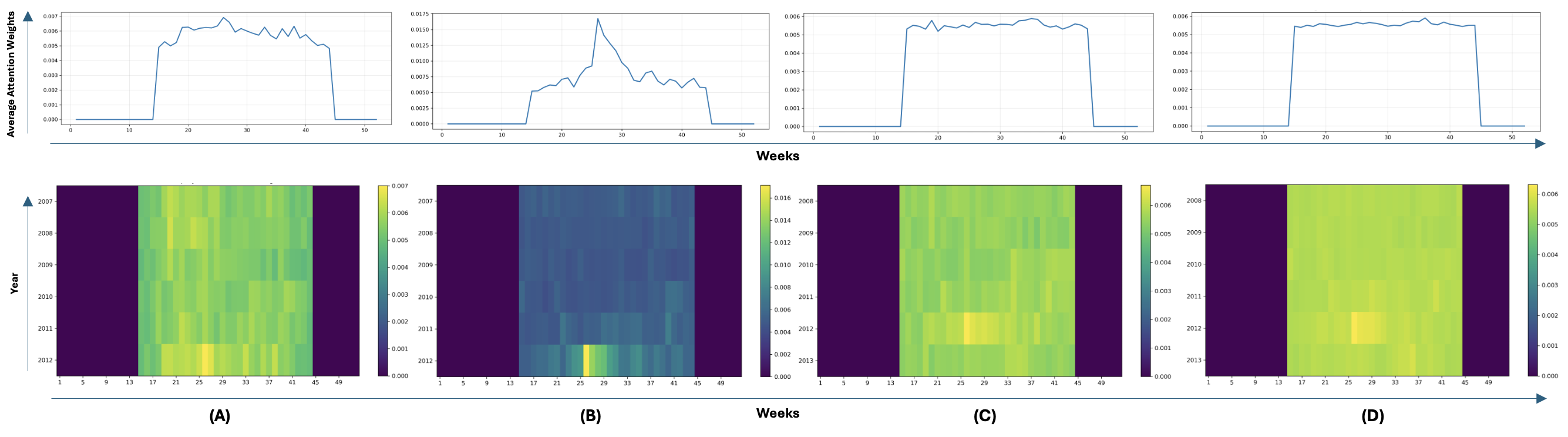}
    \caption{Capturing attention using line plots (top) and heatmaps 
    (bottom), where the heatmaps (legend: colorbar) represent the 
    average attention weights for the year. The 4 plots are for 
    (a) 2012 year with all features (VITA), (b) 2012 year with SHAP 
    features (VITA[+]), (c) 2013 year with all features (VITA), and (d) 2013 year 
    with SHAP features (VITA[+]).}
    \label{fig:attention-plot}
\end{figure*}

The SHAP-selected features are applied across both ML and deep learning models. 
{\bf However, extreme years sample weighting and attention is only applied to Tradaboost.R2 and VITA models.}

Since Tradaboost.R2 requires source and target datasets, we split the training data (data before 2012) strategically. 
The samples from the extreme event years and the previous 5 years (2007 to 2011) are considered as the target dataset, and the remaining sample from the training dataset are considered as the source dataset.
The choice of the target dataset is intentional as it allows the model to focus on samples that experience the feature distribution mismatch (extreme event years) as well as capture time-series trends from previous 5 years (since Tradaboost.R2 is a boosting based model).
We refer to this modified model as Tradaboost.R2[+].

For the VITA model, we assign higher weights to extreme years and introduce an attention bias toward the meteorological conditions of 2012's sowing and growing season; we refer to this adapted model as VITA[+].
It modifies the fine-tuning loss function by introducing a per-sample importance weight $w_i$,
\[
\mathcal{L}_{\text{VITA[+]}}
=
\frac{1}{n}
\sum_{i=1}^{n}
w_i^{\text{year}}\, (y_i - \hat{y}_i)^2
+
\beta\,
\mathrm{KL}
\!\left(
q_{\boldsymbol{\phi}}(\mathbf{z}\mid\mathbf{x})
\,\|\, 
p(\mathbf{z})
\right),
\]

where $w_i^{\text{year}} > 1$ for the identified extreme years 1983, 1988, 1993, and 2002, and $w_i^{\text{year}} = 5.0$ for the 2012 extreme year. 
To bias training toward 2012's meteorological conditions, VITA[+] computes a drift-distance using the SHAP-identified VPD feature. 
Note that ETR is excluded here as the NASA Power dataset does not contain ETR measurements mappable to the gridMET dataset.
Using 2012's sowing and growing season window (weeks 15 to 35), we construct a reference distribution ($\mu_{\text{ref}}, \sigma_{\text{ref}}$) from VPD.
For each training sample $i$, the VPD sequence is constrained to the same growing-season window, and the distance at each timestep is computed as
\[\delta_{i,t} = \left|\frac{\mathrm{VPD}_{i,t} - \mu_{\text{ref}}}{\sigma_{\text{ref}}}\right|,
\]
with the drift-distance for each sample obtained by averaging over timesteps,
\[
d_i
=
\frac{1}{|\mathcal{T}_i|}
\sum_{t \in \mathcal{T}_i}
\delta_{i,t}
\]

The base VITA model computes attention as,
\[
s_{i,t} = \mathrm{MLP}(\mathbf{z}_t),
\qquad
\alpha_{i,t} = \frac{\exp(s_{i,t})}{\sum_{k=1}^{T} \exp(s_{i,k})}.
\]

VITA[+] modifies the attention scores by subtracting the scaled 
drift-distance, $s_{i,t}^{+} = s_{i,t} - \lambda_{\text{attn}}\, d_i, \qquad \lambda_{\text{attn}} = 0.5$, yielding the adapted attention weights,
\[
\alpha_{i,t}^{+} = \frac{\exp(s_{i,t}^{+})}{\sum_{k=1}^{T} \exp(s_{i,k}^{+})}.
\]

This biases attention toward weeks whose VPD distribution most closely resembles the 2012 growing season.

\section{Evaluation}~\label{sec:evaluation}
We evaluate the model performance based on 4 performance metrics described below. 
We also elaborate on the modeling parameters for (1) All features and (2) SHAP features. 
Our results validate the model performance on extreme drought and normal year with the VITA model's attention weights explored for each year.

\subsection{Metrics}
We evaluate the performance of the machine learning models using 4 metrics: Coefficient of determination (R2), Root Mean Squared Error (RMSE), bias, and variance.
The R2 measures the variance in the observed values that the model can capture. 
It is defined as $R^2 = 1 - \frac{\sum_{i=1}^{n} (y_i - \hat{y}_i)^2}{\sum_{i=1}^{n} (y_i - \bar{y})^2}$, where $y_i$ is the observed value for the $i$-th sample and $\hat{y}_i$ is the model predictions. 
The Root Mean Squared Error (RMSE) measures the square root of the average of the squared difference between observed and predicted values as, $\mathrm{RMSE} = \sqrt{\frac{1}{n} \sum_{i=1}^{n} (y_i - \hat{y}_i)^2}$.
The bias is the mean of the prediction errors, $\text{Bias} = \frac{1}{n} \sum_{i=1}^{n} e_i$, where $e_i = y_i - \hat{y}_i$ denotes the prediction error for sample $i$.
A positive bias indicates underprediction and a negative bias indicates overprediction.
The variance is how the residuals vary around their mean and is calculated as $\text{Variance} =\frac{1}{n} \sum_{i=1}^{n} (e_i - \bar{e})^2$, where a high variance means that the model has varying performance across different observed values. 
Hence, Mean Squared Error can also be written as, $\mathrm{MSE} = \mathrm{Bias}^2 + \mathrm{Variance}$.

\subsection{Model Parameters}
The performance of the models is compared on two test scenarios, (1) the {\em extreme drought year} (2012) and (2) the {\em normal year} (2013) to compare model performance.
Each model is evaluated under two feature subsets: using all available features and SHAP features. 
{\em For Tradaboost.R2 and VITA, the SHAP feature setting in Table~\ref{tab:model_comparison_combined} corresponds to models Tradaboost.R2[+] and VITA[+] inculcated with their respective modifications as described in Section~\ref{sec:data-model-adaptation}.}

\subsubsection{Extreme drought year (2012)}
All models use standardized features and a 20-fold cross-validation.
The baseline Lasso model (Lasso 80:20) uses $\alpha = 0.5$ (All features) and $\alpha = 0.05$ (SHAP features) for its L1 penalty.
The Lasso regressor uses $\alpha = 0.005$ (All features) and $\alpha = 0.1$ (SHAP features) for its L1 penalty; Ridge regression uses $\alpha = 0.5$ (All features) and $\alpha = 500.0$ (SHAP features) for its L2 penalty. 
The Random Forest regressor uses $200$ estimators, a min sample split size of $2$, and a min sample leaf size of $1$ (All features \& SHAP features).
The XGBoost regressor uses $500$ estimators, a learning rate of $0.05$, and a mean squared error loss function (All features \& SHAP features). 
The Support Vector Regressor (SVR) uses a linear kernel where $C = 10.0$ (All features) and $C = 0.1$ (SHAP features), $\epsilon = 0.1$, and max iterations = $5000$. 
For the Tradaboost.R2, we use Ridge regressor as base estimator ($\alpha = 500.0$, $n_{estimators} = 2.0$) with the target domain having the previous 5 years from 2007 to 2011 (All features) and $\alpha = 1000.0$, $n_{estimators} = 4$) (SHAP features) with the target domain having years as $\{1983, 1988, 1993, 2002, 2007, 2008, 2009, 2010, 2011\}$.
The source domain is the remaining pre-2012 years. 
The Huber regressor uses the Huber loss $\delta = 1.35$ and L2 penalty as $\alpha = 0.01$ (All features \& SHAP features).
The Extreme Lasso regressor uses $\gamma = 3$ with $\lambda = 0.001$,  
a learning rate of $\eta = 10^{-5}$, for $2000$ iterations (All features \& SHAP features).

\subsubsection{Normal year (2013)}
For testing on the normal year, all models were similarly standardized as before and a 20-fold cross-validation was used.
The Lasso base regressor uses $\alpha = 0.05$ (All features) and $\alpha = 0.05$ (SHAP features) for its L1 penalty.
The Lasso regressor uses $\alpha = 0.05$ (All features) and $\alpha = 0.05$ (SHAP features) for its L1 penalty.
For the Tradaboost.R2, we use Lasso regressor as base estimator ($\alpha = 1.0$, $n_{estimators} = 1$) with the target domain having the previous 5 years from 2008 to 2012 (All features) and $\alpha = 1.0$, $n_{estimators} = 2$) (SHAP features) with the target domain having only normal previous years, $\{2007, 2008, 2009, 2010, 2011\}$.
The source domain is the remaining pre-2012 years. 

\subsubsection{Extreme drought \& Normal year for VITA model}
The VITA model is fine-tuned on the checkpoint using the  'small' pretrained model with the following architecture (for both 2012 and 2013 years): a Transformer encoder with $L=4$ layers, $H=10$ attention heads, hidden dimension $d=200$ ($\texttt{hidden\_dim\_factor}=20$, $d = 20 \times 10$), and feedforward dimension $4d = 800$.
The sinusoidal prior uses $k=1$ seasonal component. 
A lightweight attention pooling head (Linear($d$, 16) $\to$ GELU $\to$ Linear(16, 1)) reduces the sequence to a fixed-size representation, which is concatenated with $n_{\text{past}}=5$ prior-year yields and fed to a two-layer yield MLP (Linear($d + 6$, 120) $\to$ GELU $\to$ Linear(120, 1)).
The model is trained for 40 epochs with Adam ($\text{lr}=10^{-4}$, 10 warmup epochs), batch size 16, minimizing the ELBO with $\beta = 10^{-4}$, where $p(z)$ is the learned sinusoidal prior. 
Historical extreme years $\{1988, 1993, 2002\}$ receive a weight = $3.0$ (no 1983 since VITA pretraining starts from year 1984). 
For training, we uses prior 28 years data, with 5-fold cross validation.
{\bf The source code can be accessed at:}~\href{https://github.com/shrey-gupta/soil-land-crop}{github.com/shrey-gupta/soil-land-crop}

\subsection{Results}
\begin{table*}[htbp]
\centering
\small
\caption{Performance comparison for extreme (2012) and normal (2013) years}
\label{tab:model_comparison_combined}
\begin{tabular}{lcccccccc}
\hline
\rule{0pt}{0.20in}
 & \multicolumn{4}{c}{\textbf{All Features}} & \multicolumn{4}{c}{\textbf{SHAP Features}} \\
\cmidrule(lr){2-5} \cmidrule(lr){6-9}
\textbf{Model} 
& \textbf{$R^2$} & \textbf{RMSE} & \textbf{Bias} & \textbf{$\sqrt{\text{Var.}}$} 
& \textbf{$R^2$} & \textbf{RMSE} & \textbf{Bias} & \textbf{$\sqrt{\text{Var.}}$} \\
\hline

\multicolumn{9}{c}{\textbf{Extreme Year (2012)}} \\[0.5ex]
\hline
Lasso (80:20) & 0.788 & 19.0 & -1.96 & 18.9 & 0.808 & 18.1 & -0.09 & 18.1 \\
Lasso & 0.411 & 31.4 & 3.13 & 31.2 & 0.459 & 30.1 & -2.70 & 30.0 \\
Ridge & 0.382 & 32.1 & -7.25 & 31.3 & 0.438 & 30.6 & -6.02 & 30.0 \\
Support Vector & 0.418 & 31.2 & -4.12 & 30.9 & 0.434 & 30.8 & -6.26 & 30.1 \\
Random Forest & 0.140 & 37.9 & 19.3 & 32.6 & 0.142 & 37.9 & 18.4 & 33.1 \\
(X)GB & 0.220 & 36.1 & 17.1 & 31.8 & 0.269 & 35.0 & 14.4 & 31.8 \\
Huber & 0.430 & 30.9 & -0.403 & 30.9 & 0.440 & 30.6 & -3.34 & 30.4 \\
Tradaboost.R2 & 0.399 & 31.7 & -3.44 & 31.5 & 0.490 & 28.2 & -5.43 & 28.7 \\
Extreme Lasso & 0.045 & 40.0 & -1.75 & 39.9 & 0.059 & 39.7 & -24.4 & 31.3 \\
\rowcolor{lightgray} VITA & 0.559 & 27.1 & 4.15 & 26.8 & 0.560 & 27.1 & 1.53 & 27.1\\
\hline
\multicolumn{9}{c}{\textbf{Normal Year (2013)}} \\[0.5ex]
\hline
Lasso & 0.403 & 23.9 & -15.3 & 18.4 & 0.496 & 25.7 & -7.14 & 24.6 \\
Tradaboost.r2 & 0.356 & 24.9 & -15.9 & 19.1 & 0.421 & 23.6 & -13.9 & 19.0 \\
\rowcolor{lightgray} VITA & 0.581 & 20.0 & -5.68 & 19.2 & 0.593 & 19.7 & -6.99 & 18.5 \\
\hline
\end{tabular}
\end{table*}
We first compare all models on the extreme year 2012, and then select the two best-performing models along with the Lasso baseline for analysis on the normal year 2013. Table~\ref{tab:model_comparison_combined} reports R$^{2}$, RMSE,  bias, and variance for each model under the two feature subsets: (1) {\em All features} and (2) {\em SHAP features} (VPD and ETR).
The baseline regression model, Lasso (80:20), is trained and tested on the extreme drought year data using an 80:20 split. 
It is a useful upper-bound for the comparative models.

\subsubsection{Extreme Drought Year (2012)}
All models are trained on the years 1980 to 2011, and tested on the year 2012 as shown in Table~\ref{tab:model_comparison_combined}.
We observe that the models trained on only SHAP features show an improvement in prediction performance.
It should be noted that both VITA and VITA[+] have comparable perfromace (R2 and RMSE scores) and the VITA model, without any modifications, outperforms all competitive models. 
{\bf These results validate our questions raised in the Section~\ref{sec:introduction}: (1) VITA, a large-scale DL model, performs well under feature distribution mismatch and data irregularities unlike classical ML models.
(2) Models trained only on meteorological drivers are able to explain more than half the variations in the crop yield.
(3) Classical ML models require feature engineering and selection (SHAP) to improve their forecasting performance.}
Conventionally precipitation and temperature are considered important drivers for estimating crop yield, however, the SHAP analysis recognizes VPD and ETR as influential drivers as they include the combined effects of temperature, humidity, and radiation.
Additionally, we believe that data irregularities further suppresses the relationship between precipitation/temperature and yield as the effects of these meteorological drivers can be captured using the full annual cycle. 

\subsubsection{Normal Year (2013)}
We validate the modified models on the normal year (2013) using Lasso, TrAdaBoost.R2, and VITA models as shown in Table~\ref{tab:model_comparison_combined}.
VITA[+] has the best forecasting performance.
Unlike the extreme drought year, TrAdaBoost.R2[+] does not show an improvement over TrAdaBoost.R2 using SHAP features. 
This suggests that transfer learning models are most effective when the target and source datasets are dissimilar such as the feature distribution mismatch that exists for the 2012 yield forecasting.
Since, 2013 year does not consists of such mismatch, the underliying domain adaptation model which Tradaboost.R2 utilizes, is ineffective.

\subsubsection{Analyzing Attention for VITA}
In Figure~\ref{fig:attention-plot}, we analyze the attention weights given by the VITA model during the finetuning stage with a context window of 5 years. 
Each plot consists of a line plot representing the attention weights for the test year and a heatmap of the weights for the 5 year context window. 
In Figure~\ref{fig:attention-plot}(A) to (D), we observe that both VITA and VITA[+] models have higher attention for weeks 25 to 33, i.e., the months June to July (growing season for the corn).
This is most pronounced in Figure~\ref{fig:attention-plot}(b), where attention is almost entirely focused on these weeks. 
The fact that VITA consistently emphasizes the growing season across both 2012 and 2013 validates its attention mechanism, where it learns to prioritize growing phase months using self-supervision.

\subsection{Ablation Study}
\begin{table}[htbp]
\centering
\small
\caption{Ablation study}
\label{tab:ablation_study}
\begin{tabular}{lcccc}
\hline
\rule{0pt}{0.20in}
\textbf{Configuration} & \textbf{$R^2$} & \textbf{RMSE} & \textbf{Bias} & \textbf{$\sqrt{\text{Var.}}$} \\
\hline
\multicolumn{5}{c}{\textbf{VITA[+]}} \\
\hline
All features, no attention, weights   & 0.540 & 27.7 & 0.598 & 27.7 \\
VPD only, attention, no weights       & 0.554 & 27.3 & 4.00  & 27.0 \\
VPD only, no attention, no weights    & 0.524 & 28.2 & 5.75  & 27.6 \\
\hline
\multicolumn{5}{c}{\textbf{TradaBoost.R2[+]}} \\
\hline
All features, weights                 & 0.372 & 32.4 & -1.87 & 32.3 \\
SHAP features, no weights             & 0.478 & 29.5 & 2.65  & 29.4 \\
\hline
\end{tabular}
\end{table}
We disseminate methodological modifications of VITA[+] and Tradaboost.R2[+] using the ablation study shown in Table~\ref{tab:ablation_study}.
There are 3 variations of VITA[+] model that we use for the ablation experiments.
The first one uses \textbf{VITA[+]} model with {\bf year weights = 3.0}, {\bf all features} and {\bf attention = $0.0$} for the growing season. 
The second uses \textbf{VITA[+]} model with {\bf no year weights}, {\bf VPD feature} and {\bf attention = $0.5$} on the growing season.
The third uses \textbf{VITA[+]} model with {\bf no year weights}, {\bf VPD feature} and {\bf attention = $0.0$}.
From the VITA[+] ablation results, we observe that not having year weights, or feature attention, affects the model's performance. 
Moreover, attention on the growing season has the most effect on VITA[+]'s performance. 
It should be noted that VITA still outperforms all variations of VITA[+] (comparing Table~\ref{sec:evaluation} and Table~\ref{tab:ablation_study}), validating the fact that year weights and attention over the growing season are useful when applied together as modifications.

Similarly, there are 2 variations of the TrAdaBoost.R2[+] model in the ablation experiments shown in Table~\ref{tab:ablation_study}. 
The first uses the \textbf{TrAdaBoost.R2[+]} model with {\bf all features}, {\bf extreme year weights} and $\alpha = 5000.0$. The second uses the \textbf{TrAdaBoost.R2[+]} model with {\bf SHAP features}, {\bf no extreme year weights} and and $\alpha = 500.0$.
From the Tradaboost.R2[+] ablation results, we see that the SHAP features (VPD, ETR) play an important role for improving the performance whereas extreme year weights actually reduce the performance when tagged with all features, highlighting the importance of applying the two modifications together.

\subsection{Machine learning using 'few' test data samples}
\begin{figure}
    \centering
    \begin{subfigure}[t]{0.66\linewidth}
        \centering
        \includegraphics[width=\linewidth]{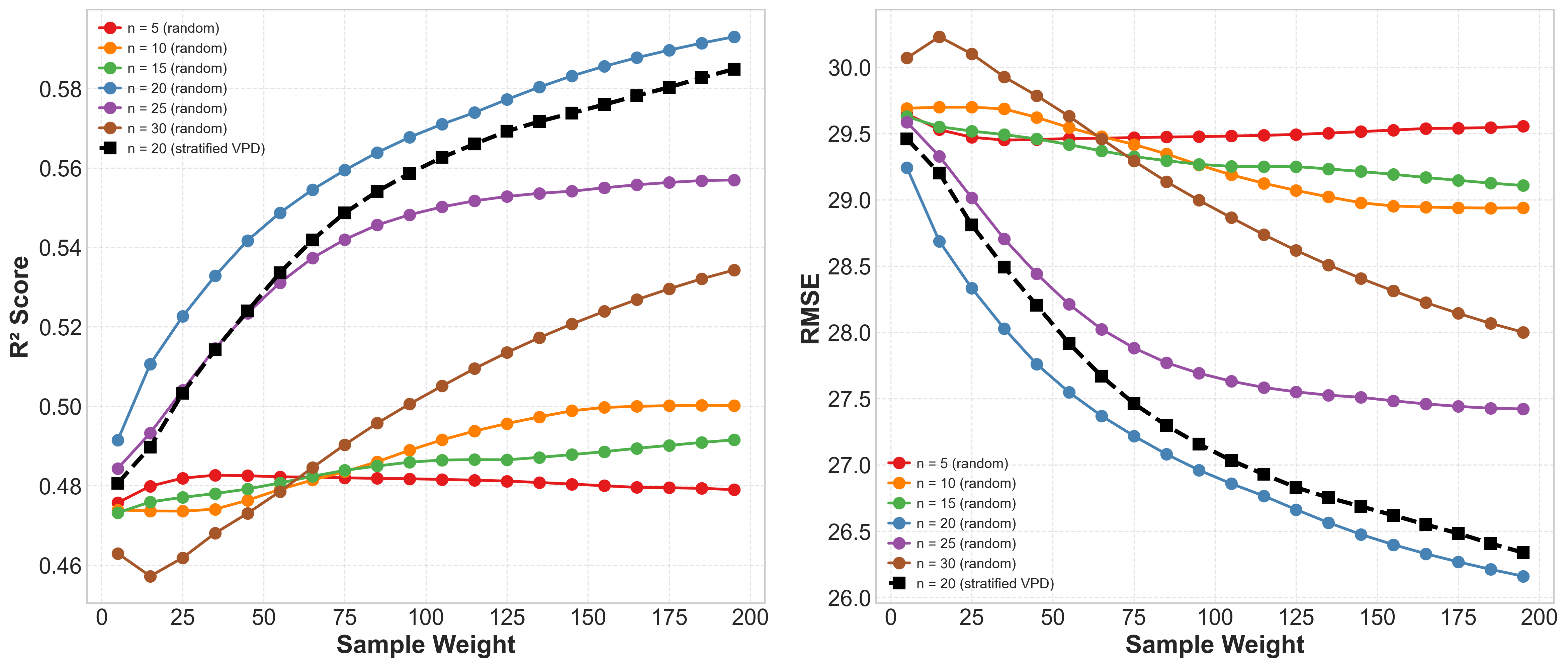}
        \caption{Plot showing performance of Lasso regressor when the samples from test data (2012 year) varies from 5 to 20 along with their weights.}
        \label{fig:few-shot-line-plot}
    \end{subfigure}
    \hfill
    \begin{subfigure}[t]{0.32\linewidth}
        \centering
        \includegraphics[width=\linewidth]{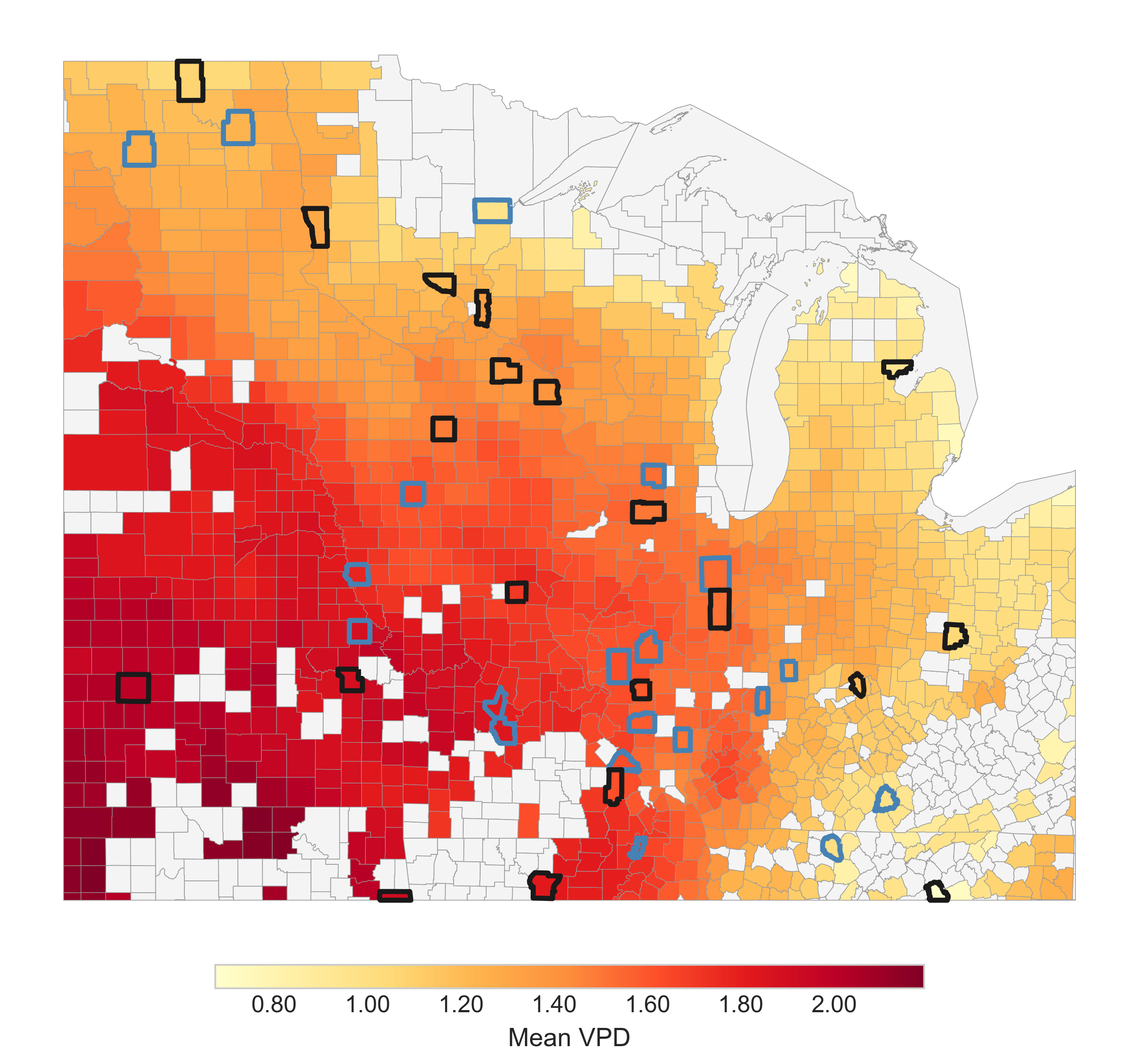}
        \caption{Map plot for the 20 counties (samples) where sampling is random v. stratified.}
        \label{fig:few-shot-map-plot}
    \end{subfigure}
    \caption{Machine learning using 'few' test data samples.}
    \label{fig:few-shot}
\end{figure}

We extend our experiments by using 'few' data samples from the extreme years as shown in Figure~\ref{fig:few-shot}. 
We use $5$ to $30$ samples selected randomly where weights varying from $5$ to $200$ is assigned to each of them and the Lasso regressor with $alpha=0.05$ is used.
In Figure~\ref{fig:few-shot-line-plot}, we observe that for $n=20$, the model has the best performance which improves with the weights for the samples.
We also plot these 20 samples (counties) in Figure~\ref{fig:few-shot-map-plot}, where counties highlighted in blue are the randomly selected 20 samples.
Based on these results, we performed stratified sampling for $n=20$ data samples (counties). 
To ensure an equal coverage, we first sorted the counties based on the mean of the growing-season (April–August) of vapor pressure deficit (VPD).
Then we selected $20$ counties via uniform sampling across the sorted counties. 
The results of this sampling is present in Figure~\ref{fig:few-shot-line-plot} and the (black) highlighted counties in Figure~\ref{fig:few-shot-map-plot}.
While stratification can be performed in numerous ways such as selecting large number of VPD stress counties, our motivation was to improve generalizability.
We observe that the model with stratified sampling underperform compared to random sampling in Figure~\ref{fig:few-shot-line-plot} indicating that yield variability cannot be solely determined using VPD stress on crops.

These experiments give insights into the advantages of having 'few' labeled county data for the extreme drought year.
While having meteorological data from the growing season is possible for forecasting the yield variability, there still remains the question of knowing the optimal number of 'few' test samples ($n$ ranging from $5$ to $20$) that requires the labeled test data itself.

\section{Discussion and Future Work}~\label{sec:discussion}
\begin{figure*}[t]
    \centering
    \begin{subfigure}[t]{0.61\textwidth}
        \centering
        \includegraphics[width=\linewidth]{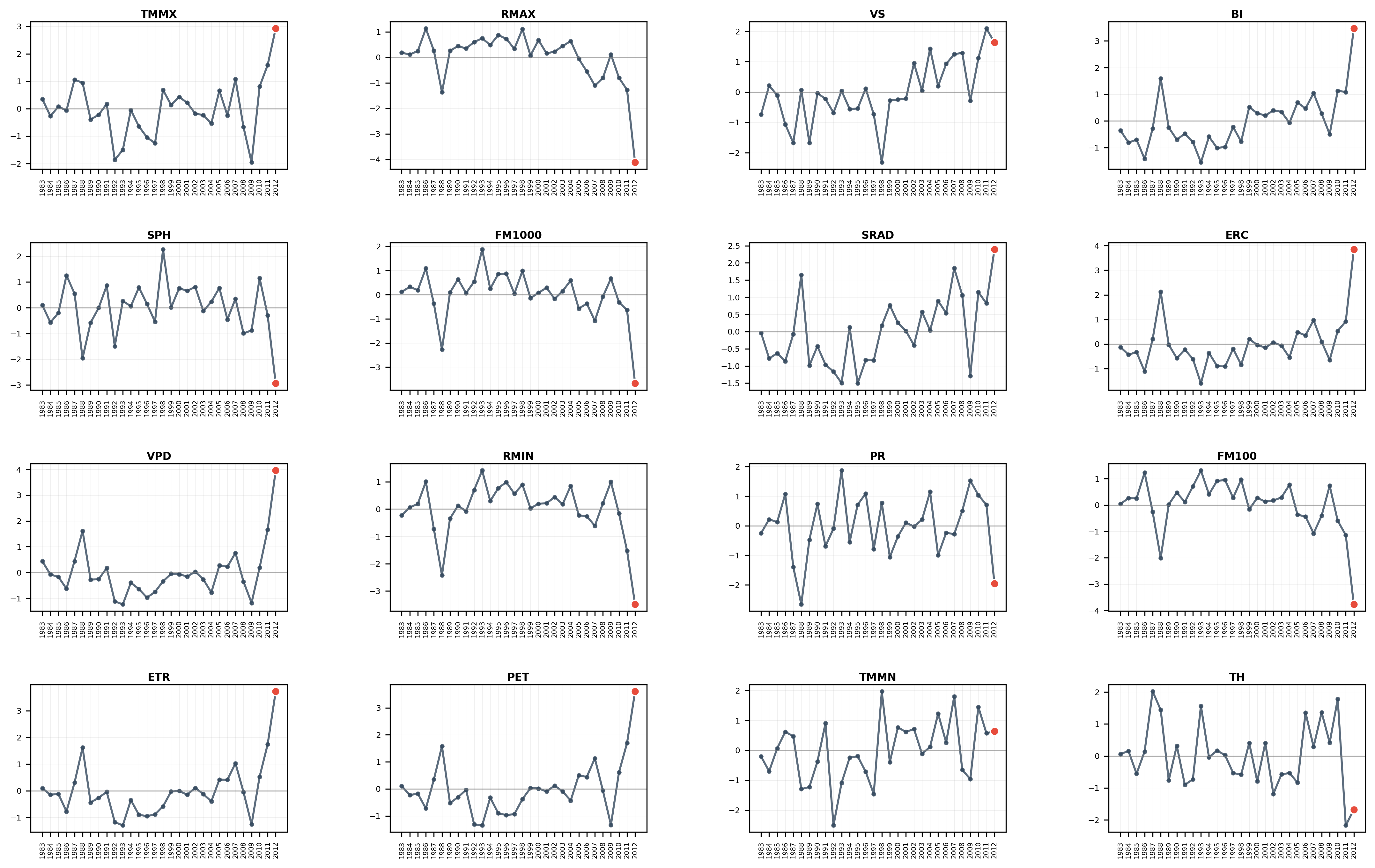}
        \caption{Annual aggregated values of features across all counties}
        \label{fig:annual-aggregated}
    \end{subfigure}
    \hfill
    \begin{subfigure}[t]{0.38\textwidth}
        \centering
        \includegraphics[width=\linewidth]{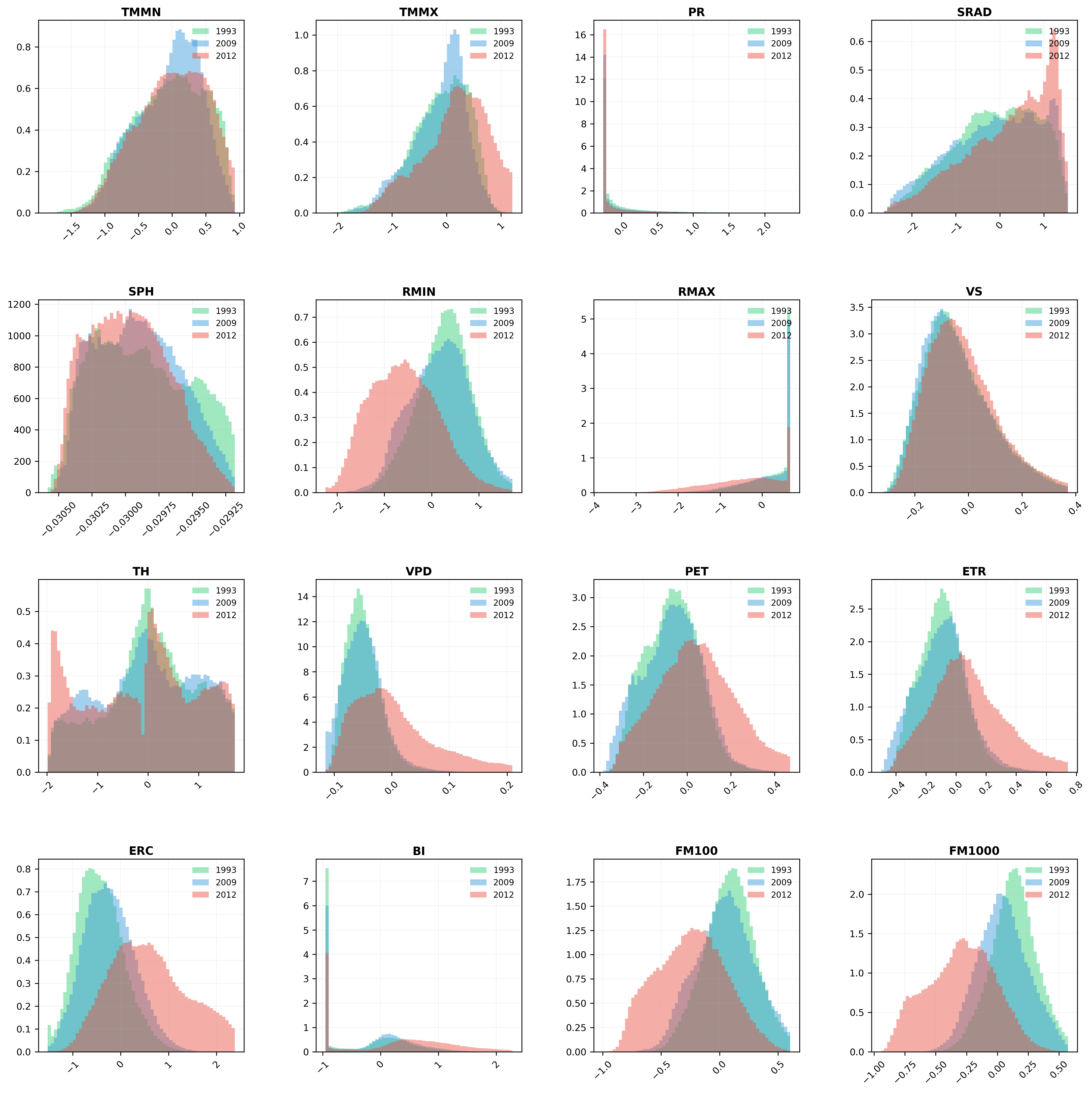}
        \caption{Density plots for extreme and normal years}
        \label{fig:denisty-plot}
    \end{subfigure}

    \caption{Measuring feature distribution mismatch for the extreme event years using (a) Aggregated annual feature values across all counties over the years, (b) Density plot of daily values for extreme (2012 and 1993) and normal (2009) years across all counties.}
    \label{fig:feature distribution mismatch}
\end{figure*}

The gridMET dataset has spatiotemporal irregularities that are common across environmental and climate science domains and consequently shapes the goal of this study. 
Since our models are adapted to work with feature distribution mismatch scenarios, we first discuss how changes in data distribution is observed for the extreme drought year.
We then discuss the data and modeling extensions that the future work can entail.

\paragraph{{\bf Feature distribution mismatch}}
As defined previously, feature distribution mismatch is observed when the feature distribution of the training and test datasets differ. 
In Figure~\ref{fig:feature distribution mismatch}, we observe how the feature values change over the years where Figure~\ref{fig:feature distribution mismatch}(a) shows the number of standard deviations by which the aggregated feature value (aggregated across days and counties) deviates from the mean across all the years.
We use the z-scores to measure deviations from the mean and observe that 2012 show large deviations, especially for the SHAP features VPD (Vapor Pressure Deficit) and ETR (Evapotranspiration). 
Moreover, not all features show such deviations from the mean like Precipitation (PR) and Minimum temperature (TMMN) which are less skewed compared to their observed values for the years 1992 and 1998.

We observe a similar trend in the density plot shown in Figure~\ref{fig:feature distribution mismatch}(b) which calculates the standard deviation for a feature across all counties in a given year.
We compare three years -- 1993 (extreme drought year), 2009 (normal year) and 2012 (extreme drought year).
Then we bin them using 60 bins and clip values more than the 99th percentile to avoid long tails and improve visual representation.
We observe that VPD and ETR show a heavy tail compared to the normal year (2009) and the extreme drought year (1993).
While a deviation from the normal year is expected, it is interesting to see the deviation of 2012 from an earlier extreme drought year, 1993. 
This also validate the claim that the 2012 year has samples with more extreme values than observed historically.

\paragraph{{\bf Overcoming Data Scarcity}}
Data availability and granularity are central to effective modeling. Even robust models can underperform when training data is scarce, though open-source datasets and collaborative efforts can mitigate unnecessary constraints. 
Here we discuss the data limitations of the current study and outline directions for addressing them.
\begin{enumerate}
    \item {\em Removing spatiotemporal irregularities} The gridmet dataset has spatiotemporal irregularities where only 210/365 days of daily data is used, which might not be able to capture long-term precipitation and temperature trends that contribute to the variations in the yield.
    A study that employs complete data will be useful to capture these trends.
    \item {\em Improving resolution for the pretrained model} The deep learning model, VITA, relies on the coarse NASA Power dataset with a lower resolution of $0.5^\circ$. 
    A higher resolution grid data using climate models such as GFDL AM4~\cite{zhao2018gfdl,horowitz2020gfdl} or the ERA5 dataset can be used here to improve the pretraining accuracy of the model.
    \item {\em Increasing granularity for the region of cover} Forecasting yields for a larger region (counties) as compared to smaller areas, ranging within the 5x5 km scale, affects the forecasting results. 
    A more local forecasting within a smaller region will be useful to test VITA's learning and performance.
\end{enumerate}

\section{Conclusion}
In this study, we observe that large-scale models such as VITA are highly effective for forecasting crop yield under feature distribution mismatch scenarios like the 2012 drought in the US Corn belt. 
Moreover, with or without modifications, which we call model adaptation for the data, VITA consistently outperforms machine learning (non-deep learning) models.
This indicates that pretrained deep learning models like VITA are able to overcome feature distribution mismatch, and spatiotemporal irregularities in the dataset, achieving strong prediction performance from meteorological drivers alone.
However, the model adaptation using SHAP feature selection and extreme-year weighting shows that the performance of ML models can also be improved for this scenario. 
Overall, this study suggest that ML explainability based modifications are most useful when large-scale models cannot be deployed.
Whereas, deep learning models performance, despite their black box nature, improves as data quality and quality increases (pretrained v. non-pretrained deep learning models).

\end{document}